\documentclass{article}
\usepackage{spconf,amsmath,graphicx}
\usepackage[colorlinks=false]{hyperref}
\usepackage{placeins}
\usepackage{threeparttable}
\usepackage{multirow}
\usepackage{balance}

\title{Self-Trained Model for ECG Complex Delineation}

\name{Aram Avetisyan$^1$, Nikolas Khachaturov$^2$, Ariana Asatryan$^3$, Shahane Tigranyan$^3$, Yury Markin$^1$}

\address{$^1$Institute for System Programming of the Russian Academy of Sciences, Moscow, Russia\\
$^2$Moscow State University, Yerevan, Armenia\\
$^3$Russian-Armenian University, Yerevan, Armenia\\
}

\begin{document}

\maketitle

\begin{abstract}

Electrocardiogram (ECG) delineation plays a crucial role in assisting cardiologists with accurate diagnoses. Prior research studies have explored various methods, including the application of deep learning techniques, to achieve precise delineation. However, existing approaches face limitations primarily related to dataset size and robustness.

In this paper, we introduce a dataset for ECG delineation and propose a novel self-trained method aimed at leveraging a vast amount of unlabeled ECG data. Our approach involves the pseudolabeling of unlabeled data using a neural network trained on our dataset. Subsequently, we train the model on the newly labeled samples to enhance the quality of delineation. We conduct experiments demonstrating that our dataset is a valuable resource for training robust models and that our proposed self-trained method improves the prediction quality of ECG delineation.

\end{abstract}


\section{Introduction}
\label{sec:intro}

\begin{figure}[]
\includegraphics[width=0.7\columnwidth]{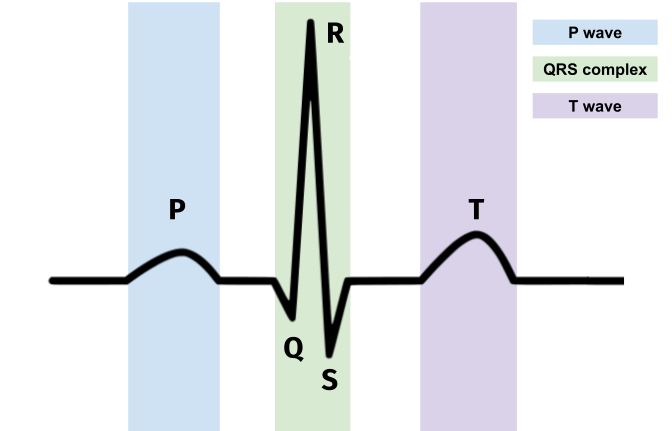}
\centering
\caption{PQRST segment of an ECG signal.}
\label{fig:PQRST}
\end{figure}

The analysis of electrocardiograms (ECGs) continues to be one of the key instruments for assessing patients' cardiac health. With the exponential growth in the volume of digital electrocardiograms, new opportunities have arisen for automatically diagnosing heart abnormalities and other ECG-related tasks~\cite{hong2020opportunities,liu2021deep}. Among the critical tasks in ECG analysis is precise ECG record delineation, as it can help clinicians in making accurate diagnoses. Figure~\ref{fig:PQRST} demonstrates the example of delineated fiducial points within the PQRST complex that constitute the record.
\let\thefootnote\relax\footnotetext{
Email: a.a.avetisyan@ispras.ru, nickhach23@gmail.com,\\ arianasatryan@gmail.com,shahane.tigranyan99@gmail.com,ustas@ispras.ru}

Early studies in the field of ECG delineation analysis primarily relied on conventional mathematical techniques, with a predominant focus on QRS complexes~\cite{pan1985real,nygaards1983delineation}. Subsequent advancements led to the development of new delineation techniques, most notably those based on wavelet transforms, designed to comprehensively capture ECG features, including the P and T waves of the signal~\cite{li1995detection,martinez2004wavelet}.

With the advent of datasets~\cite{kalyakulina2020ludb,laguna1997database} with fiducial point labels, researchers have embraced the potential of neural networks to achieve more effective ECG delineation. As a natural progression, deep learning methods have been applied to ECG analysis, using annotated complexes to improve accuracy. These deep learning methods typically adopt architectures such as encoder-decoder structures~\cite{moskalenko2020deep,liang2022ecg_segnet} or LSTM-based models~\cite{peimankar2021dens,abrishami2018supervised} often incorporating additional domain knowledge~\cite{wang2020knowledge} to improve prediction quality.

However, despite advancements in fiducial point detection, existing datasets remain relatively small. While research in ECG abnormality classification benefits from extensive data utilization~\cite{avetisyan2024deep,ribeiro2020automatic}, studies in ECG delineation face challenges in achieving higher generalization capabilities. Datasets often lack annotations and fail to adequately represent the diversity of cardiac abnormalities. Moreover, open datasets are typically characterized by low noise levels, potentially causing models to perform poorly when exposed to data with higher levels of noise.

In this study, we address these challenges by introducing a novel dataset of ECG records, meticulously annotated with fiducial points. We present the ECG-CODE model and propose the self-trained method to enhance model prediction. We conduct experiments on both our collected dataset and the existing open dataset and demonstrate the effectiveness of our approach.

\section{Methods}
\label{sec:pagestyle}

\subsection{Dataset Description}
\label{sec:data_description}

Our study leverages two distinct datasets for the ECG PQRST delineation task: the publicly available LUDB dataset~\cite{kalyakulina2020ludb} and the collected ISP dataset meticulously annotated by two cardiologists. The ISP dataset is a collection of 530 labeled 10-second 12-lead ECG records, which contains three times more fiducial points compared to all the open datasets for ECG delineation. The ISP dataset offers a rich variety of cardiac abnormalities and has undergone thorough annotation by 2 cardiologists.\let\thefootnote\relax\footnotetext{The dataset is available at 10.5281/zenodo.11472365} We do not include the second popular open dataset QT~\cite{laguna1997database} in our experiments, because it contains two-lead recordings, whereas we use 12-lead ECG records in our study. We provide a comprehensive comparative analysis of annotated waveforms in the datasets used in the work, as summarized in Table~\ref{table:characteristics}.

For pseudolabeling purposes, we harnessed the PTB-XL dataset~\cite{wagner2020ptb}, an open dataset comprising more than 20000 12-lead ECGs. Unlike the described datasets, PTB-XL lacks fiducial point annotations, yet it offers an invaluable resource for our model's relabeling approach.

\begin{table}[!hb]
\caption{Characteristics of ISP and LUDB datasets.}
\label{table:characteristics}
\centering
\begin{tabular}{|c|c|c|c|c|c|}
\hline
& \textbf{Records} & \textbf{P wave}  & \textbf{QRS complex} & \textbf{T wave}  \\ \hline
ISP & 499 & 5,179 & 6,284 & 5,768 \\
LUDB & 200 & 1,393 & 1,803 & 1,621 \\ \hline
\end{tabular}
\end{table}

\subsection{Methodology}
\label{sec:network_structure_explanation}

\begin{table*}[]
\caption{Delineation performance of ECG-CODE models trained on ISP and LUDB.}
\label{table:First experiment}
\centering
\resizebox{0.8\textwidth}{!}{
\begin{threeparttable}
    \begin{tabular}{|c|c|c|c|c|c|c|c|}
        \hline
\multicolumn{2}{|c|}{\textbf{}} & \multicolumn{6}{|c|}{\textbf{Tested on LUDB}} \\ \hline
\multicolumn{2}{|c|}{\textbf{}} & \textbf{P on} & \textbf{P off} & \textbf{QRS on} & \textbf{QRS off} & \textbf{T on} & \textbf{T off} \\ \hline 
\multirow{4}{*}{\textbf{Trained on LUDB}} & \textit{Se} & 0.950 & 0.950 & 0.997 & 0.997 & 0.986 & 0.954 \\ & \textit{PPV} & 0.986 & 0.986 & 1.000 & 1.000 & 1.000 & 0.968 \\ & \textit{F1-score} & 0.968 & 0.968 & 0.999 & 0.999 & \textbf{0.993} & 0.961 \\ & $\mu \pm \sigma$ & \(-5.1 \pm 20.6\) & \(5.1 \pm 17.5\) & \(3.6 \pm 10.5\) & \(2.4 \pm 16.4\) & \(-5.5 \pm 32.5\) & \(5.0 \pm 24.8\) \\ \hline
\multirow{4}{*}{\textbf{Trained on ISP}} & \textit{Se} & 0.983 & 0.983 & 1.000 & 1.000 & 0.983 & 0.954 \\ & \textit{PPV} & 0.967 & 0.977 & 1.000 & 1.000 & 0.997 & 0.968 \\ & \textit{F1-score} & \textbf{0.975} & \textbf{0.985} & \textbf{1.000} & \textbf{1.000} & 0.990 & \textbf{0.961} \\ & $\mu \pm \sigma$ & \(-1.2 \pm 20.3\) & \(-7.5 \pm 14.2\) & \(13.9 \pm 10.3\) & \(-12.0 \pm 16.3\) & \(13.7 \pm 32.5\) & \(-7.5 \pm 28.1\) \\ \hline
\multicolumn{2}{|c|}{\textbf{}} & \multicolumn{6}{|c|}{\textbf{Tested on ISP}} \\ \hline
\multicolumn{2}{|c|}{\textbf{}} & \textbf{P on} & \textbf{P off} & \textbf{QRS on} & \textbf{QRS off} & \textbf{T on} & \textbf{T off} \\ \hline
\multirow{4}{*}{\textbf{Trained on LUDB}} & \textit{Se} & 0.813 & 0.818 & 0.957 & 0.957 & 0.888 & 0.865 \\ & \textit{PPV} & 0.949 & 0.955 & 0.992 & 0.992 & 0.954 & 0.930 \\ & \textit{F1-score} & 0.876 & 0.881 & 0.974 & 0.974 & 0.920 & 0.896 \\ & $\mu \pm \sigma$ & \(-6.7 \pm 22.9\) & \(11.0 \pm 15.7\) & \(-6.3 \pm 11.2\) & \(16.5 \pm 15.6\) & \(-16.2 \pm 33.7\) & \(11.5 \pm 29.6\) \\ \hline
\multirow{4}{*}{\textbf{Trained on ISP}} & \textit{Se} & 0.967 & 0.967 & 0.987 & 0.987 & 0.951 & 0.938 \\ & \textit{PPV} & 0.956 & 0.956 & 0.998 & 0.998 & 0.955 & 0.942 \\ & \textit{F1-score} & \textbf{0.961} & \textbf{0.961} & \textbf{0.993} & \textbf{0.993} & \textbf{0.953} & \textbf{0.940} \\ & $\mu \pm \sigma$ & \(-0.9 \pm 17.9\) & \(-3.1 \pm 14.3\) & \(4.1 \pm 10.0\) & \(-1.6 \pm 12.4\) & \(4.1 \pm 29.4\) & \(1.0 \pm 26.6\) \\ \hline
    \end{tabular} 
\end{threeparttable}}
\end{table*}

\begin{figure}[]
\includegraphics[width=\columnwidth]{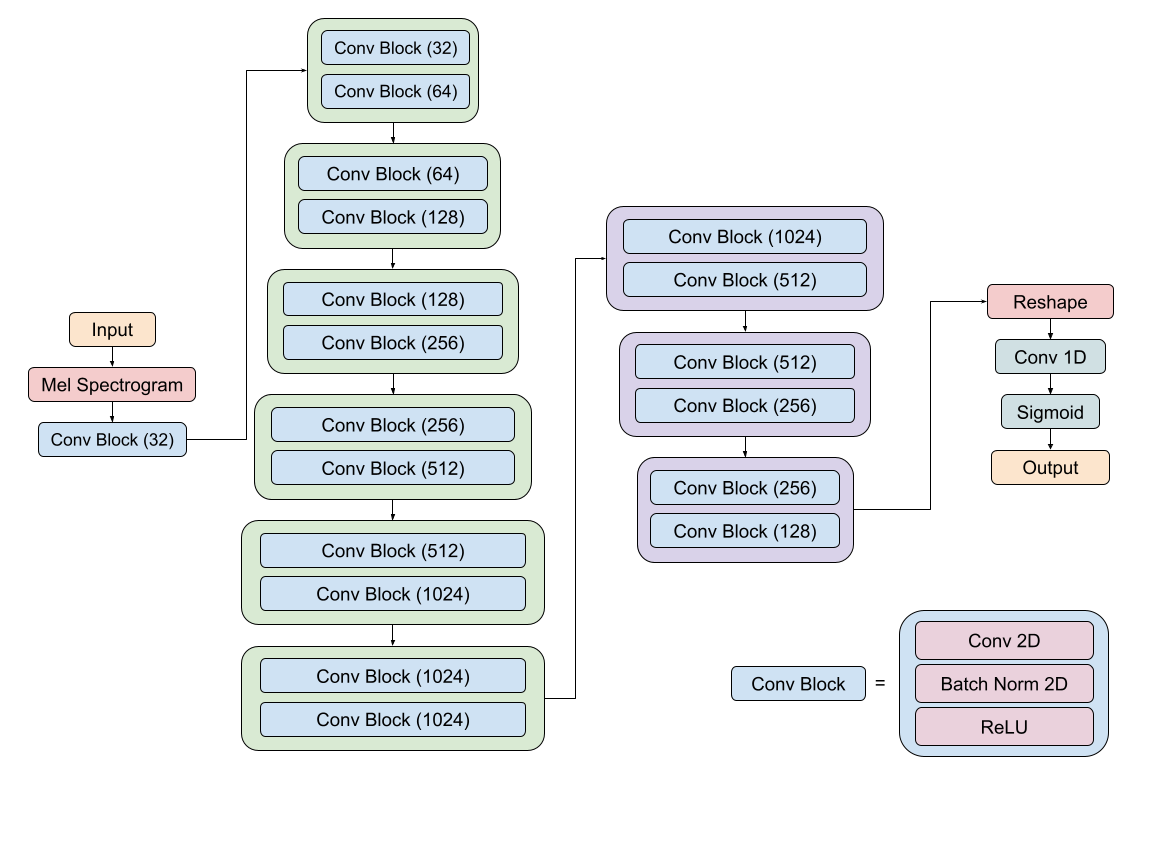}
\centering
\caption{The architecture of ECG-CODE network.}
\label{fig:ECG_CODE}
\end{figure}

We propose a Convolutional Neural Network-based model for ECG COmplex DElineation (ECG-CODE). The network can use various number of leads of the signal as an input. In our study we used all 12 leads of an ECG record as an input and transformed them into mel spectrograms. Within the ECG-CODE model, we incorporate MobileNet block structures~\cite{howard2017mobilenets} to modify dimensionality, enabling systematic expansion and compression of the information space. A detailed view of the ECG-CODE model architecture is demonstrated at Figure~\ref{fig:ECG_CODE}.

For the output the record is divided into multiple non-overlapping intervals and the model outputs prediction for each of these intervals. Information about each interval contains a confidence score and start-end points for each segment the model predicts: P-wave, QRS, and T-wave. The model's objective is to predict, for each of the three considered objects (P-wave, QRS, T-wave), whether the object is present within an interval, and if it is, to predict its start and end.

During training, we use a specific loss function that consists of confidence loss (CL) and start-end loss (SEL). CL is responsible for correctly predicting the presence of an object within the interval. SEL penalizes the model for errors in predicting fiducial points for intervals in which the object is present. The formulas for CL and SEL are the following:

\begin{equation}
\footnotesize
CL =
\begin{cases}
    0, & \text{if } |PC - TC| < 0.25 \\
    (PC - TC)^{2}, & \text{if } |PC - TC| \geq 0.25
\end{cases}
\end{equation}

\begin{equation}
\footnotesize
SS = (PS - TS)^{2} + (PE - TE)^{2}
\end{equation}

\begin{equation}
\footnotesize
SEL =
\begin{cases}
    0, & \text{if } SS < 0.15 \\
    SS \cdot TC, & \text{if } SS \geq 0.15
\end{cases}
\end{equation}

where TC and PC are the target confidences and prediction confidences, PS and TS are prediction and target starts, PE and TE are prediction and target ends respectively. The resulting loss of the model is taken as the sum of CL and SEL.

After the prediction, we applied post-processing to the output. In the first stage, we combined all pairs of segments of one object with a distance between them less than a predefined number of dots. Subsequently, we removed all segments less than a specified minimum length.

To use the advantage of a substantial volume of unlabeled data, we propose a novel ECG self-trained approach. We employ a pre-trained ECG-CODE model to predict segments within the unlabeled dataset. For each record in the dataset, we extract class labels, their respective locations on the record, and the associated prediction confidence.

To enhance the reliability of these predictions, we sort all predicted records by delineation score, separately for each object. To calculate the delineation score of the prediction we use the following formula:

\begin{equation}
\footnotesize
    \textnormal{delineation score} = |0.5 - \textnormal{predicted confidence}|
\end{equation}

Each predicted confidence has a value in the range $(0, 1)$ and the delineation score ensures that we will consider the prediction of both the presence and absence of objects in the intervals equally. As a result, we take the mean and standard deviation of predicted confidences for each wave: P-wave, QRS, and T-wave.

Using the resulting confidence scores, we create a new dataset by selecting the top N\% of predictions for each object, with N as a hyperparameter. The new model is trained from scratch using this newly labeled data and further fine-tuned on the manually annotated dataset.

\subsection{Implementation Details}
\label{sec:implementation_details}

\begin{figure*}[t]
\includegraphics[width=0.7\textwidth]{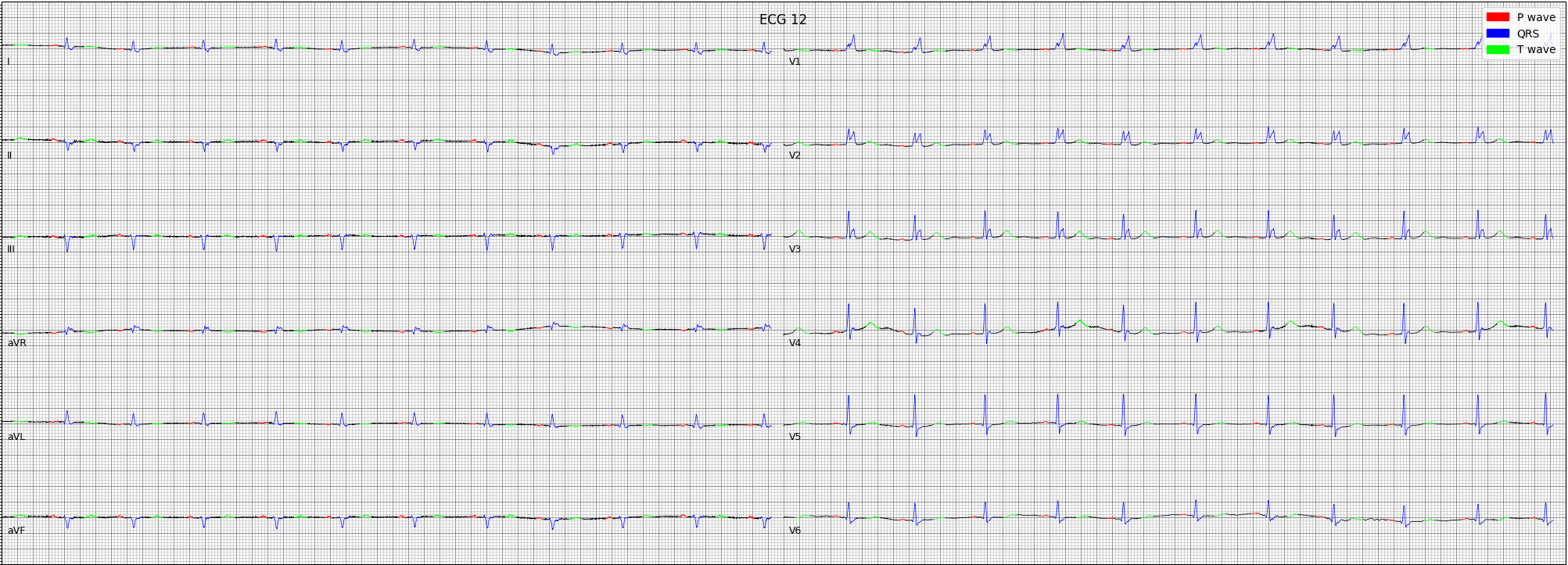}
\centering
\caption{Example of ECG signal delineated by ECG-CODE trained from scratch of pseudolabeled PTB-XL and fine-tuned on ISP.}
\label{fig:segmentation}
\end{figure*}

We trained the ECG-CODE on the ISP training dataset to get the initial model for ECG delineation. Then, we used the PTB-XL dataset for self-trained and fine-tuning ECG-CODE. The proportion of PTB-XL for pseudolabeling was chosen to be 50\%. We applied data augmentation during training, which involved sequential operations such as the Bandpass filter and Notch filter, each with a 0.5 probability of application. Finally, we resampled all records to a 1000 Hz sampling frequency and applied z-score normalization to ensure uniformity. In the postprocessing phase, we set the maximum length to unite segments at 300 dots and the minimum length for a segment at 50 dots.

To evaluate the quality of our models' predictions, we compared the predicted fiducial points to the ground truth points. The model was designed to predict six possible fiducial points: the onset and offset of the P wave (P\_onset, P\_offset), the onset and offset of the QRS segment (QRS\_onset, QRS\_offset), and the onset and offset of the T wave (T\_onset, T\_offset).

We evaluated each predicted point by determining whether it fell within a predefined region, we named window tolerance. Predictions within this window tolerance of the corresponding ground truth points with matching classes were categorized as True Positives. Predictions without any corresponding ground truth points within the window tolerance were considered False Positives. Instances where ground truth points failed to fall within any window tolerance region were marked as False Negatives.

Furthermore, we quantified the accuracy of our predictions by summarizing the errors for each True Positive target. As a result, we evaluated the model's performance using a range of metrics, including sensitivity (Se), positive predictive value (PPV), F1-score, as well as the mean and standard deviation of errors measured in milliseconds.

\section{Experiments and results}
\label{sec:typestyle}

\begin{table*}[]
\caption{Delineation performance of ECG-CODE model trained on ISP dataset (ECG-CODE) and ECG-CODE model trained from scratch of pseudolabeled PTB-XL and fine-tuned on ISP.}
\label{table:second experiment}
\centering
\resizebox{0.8\textwidth}{!}{
\begin{threeparttable}
    \begin{tabular}{|c|c|c|c|c|c|c|c|}
        \hline
\multicolumn{2}{|c|}{\textbf{}} & \textbf{P on} & \textbf{P off} & \textbf{QRS on} & \textbf{QRS off} & \textbf{T on} & \textbf{T off} \\ \hline
\multirow{4}{*}{\textbf{ECG-CODE}} & \textit{Se} & 0.967 & 0.967 & 0.987 & 0.987 & 0.951 & 0.938 \\ & \textit{PPV} & 0.956 & 0.956 & 0.998 & 0.998 & 0.955 & 0.942 \\ & \textit{F1-score} & 0.961 & 0.961 & \textbf{0.993} & \textbf{0.993} & 0.953 & 0.940 \\ & $\mu \pm \sigma$ & \(0.9 \pm 17.9\) & \(-3.1 \pm 14.3\) & \(4.1 \pm 10.0\) & \(-1.6 \pm 12.4\) & \(4.1 \pm 29.4\) & \(1.0 \pm 26.6\) \\ \hline
\multirow{4}{*}{\textbf{Pseudolabeled ECG-CODE }} & \textit{Se} & 0.957 & 0.959 & 0.983 & 0.983 & 0.958 & 0.950 \\ & \textit{PPV} & 0.979 & 0.981 & 0.996 & 0.996 & 0.964 & 0.956 \\ & \textit{F1-score} & \textbf{0.968} & \textbf{0.970} & 0.990 & 0.990 & \textbf{0.961} & \textbf{0.953} \\ & $\mu \pm \sigma$ & \(1.1 \pm 16.5\) & \(-0.2 \pm 13.6\) & \(2.6 \pm 9.9\) & \(-0.3 \pm 11.5\) & \(-0.1 \pm 27.5\) & \(2.5 \pm 23.3\) \\ \hline
    \end{tabular}
\end{threeparttable}}
\end{table*}

\subsection{Experimental setup}
\label{ssec:subhead}

In this study, we conducted a series of experiments to validate the effectiveness of our proposed methodology.

In the first experiment, we aimed to comparee the performance of the ECG-CODE model trained on two different datasets: LUDB and ISP. Both models were evaluated on the test sets of the observed datasets. We maintained the same evaluation parameters of these methods to ensure a fair comparison and tested the models with a window tolerance of 150 milliseconds. Notably, we excluded the first and last seconds of records in the LUDB dataset, as these segments lacked annotations.

In the second experiment, our focus shifted to assessing the impact of self-trained on prediction quality. To achieve this, we compared the performance of ECG-CODE model trained on ISP dataset (ECG-CODE) against an ECG-CODE model trained from scratch of pseudolabeled PTB-XL and fine-tuned on ISP (Pseudolabeled ECG-CODE). We conducted evaluations on ISP test dataset. Similar to the first experiment, we set a fixed window tolerance of 150 milliseconds, sufficient for reliable prediction estimates.

All evaluations were performed on raw test data to maintain the integrity of the experiment. The f1-score metric was selected as the key metric for comparison.

\subsection{Experiment results}
\label{ssec:subhead}

The results of the first experiment are summarized in Table~\ref{table:First experiment}. Notably, the model trained on the LUDB dataset got significantly higher metrics when tested on its own samples compared to the ISP test set. Conversely, ECG-CODE trained on the ISP data demonstrated stable performance on both datasets, and even better performance for several points on LUDB samples than a network trained on LUDB. When evaluated on the ISP data, the results of the model trained on LUDB decreased by 10\% and 4\% for P and T waves respectively. The results highlights the effectiveness of the model and the advantage of the proposed ISP dataset, demonstrating that training on ISP data yields more robust and generalized models.

The results of the second experiment are presented in Table~\ref{table:second experiment}. The model that incorporated self-trained during training achieved the highest quality. In addition, it is worth noting that even with limited data, the model almost perfectly copes with the task of isolating the QRS complex. Therefore, the primary contribution of self-trained lies in its ability to identify P and T waves, critical for determining various cardiac abnormalities.

Additionally, in Figure~\ref{fig:segmentation} we provide an example of delineation performed by Pseudolabeled ECG-CODE model.

\section{Conclusion}
\label{sec:page}

In this study, we introduced the ISP dataset and proposed an innovative self-trained approach for ECG delineation. The introduction of the ISP enriches the landscape of ECG delineation datasets, offering a valuable resource for training robust models. Our self-trained method demonstrates its effectiveness in improving the quality of fiducial points prediction within an ECG record.

\vfill\pagebreak


\begin{thebibliography}{10}

\bibitem{hong2020opportunities}
Shenda Hong, Yuxi Zhou, Junyuan Shang, Cao Xiao, and Jimeng Sun,
\newblock ``Opportunities and challenges of deep learning methods for
  electrocardiogram data: A systematic review,''
\newblock {\em Computers in biology and medicine}, vol. 122, pp. 103801, 2020.

\bibitem{liu2021deep}
Xinwen Liu, Huan Wang, Zongjin Li, and Lang Qin,
\newblock ``Deep learning in ecg diagnosis: A review,''
\newblock {\em Knowledge-Based Systems}, vol. 227, pp. 107187, 2021.

\bibitem{pan1985real}
Jiapu Pan and Willis~J Tompkins,
\newblock ``A real-time qrs detection algorithm,''
\newblock {\em IEEE transactions on biomedical engineering}, , no. 3, pp.
  230--236, 1985.

\bibitem{nygaards1983delineation}
M~E Nyg{\aa}rds and L~S{\"o}rnmo,
\newblock ``Delineation of the qrs complex using the envelope of the ecg,''
\newblock {\em Medical and biological engineering and computing}, vol. 21, pp.
  538--547, 1983.
\balance
\bibitem{li1995detection}
Cuiwei Li, Chongxun Zheng, and Changfeng Tai,
\newblock ``Detection of ecg characteristic points using wavelet transforms,''
\newblock {\em IEEE Transactions on biomedical Engineering}, vol. 42, no. 1,
  pp. 21--28, 1995.

\bibitem{martinez2004wavelet}
Juan~Pablo Mart{\'\i}nez, Rute Almeida, Salvador Olmos, Ana~Paula Rocha, and
  Pablo Laguna,
\newblock ``A wavelet-based ecg delineator: evaluation on standard databases,''
\newblock {\em IEEE Transactions on biomedical engineering}, vol. 51, no. 4,
  pp. 570--581, 2004.

\bibitem{kalyakulina2020ludb}
Alena~I Kalyakulina, Igor~I Yusipov, Viktor~A Moskalenko, Alexander~V
  Nikolskiy, Konstantin~A Kosonogov, Grigory~V Osipov, Nikolai~Yu Zolotykh, and
  Mikhail~V Ivanchenko,
\newblock ``Ludb: a new open-access validation tool for electrocardiogram
  delineation algorithms,''
\newblock {\em IEEE access}, vol. 8, pp. 186181--186190, 2020.

\bibitem{laguna1997database}
Pablo Laguna, Roger~G Mark, A~Goldberg, and George~B Moody,
\newblock ``A database for evaluation of algorithms for measurement of qt and
  other waveform intervals in the ecg,''
\newblock in {\em Computers in cardiology 1997}. IEEE, 1997, pp. 673--676.

\bibitem{moskalenko2020deep}
Viktor Moskalenko, Nikolai Zolotykh, and Grigory Osipov,
\newblock ``Deep learning for ecg segmentation,''
\newblock in {\em Advances in Neural Computation, Machine Learning, and
  Cognitive Research III: Selected Papers from the XXI International Conference
  on Neuroinformatics, October 7-11, 2019, Dolgoprudny, Moscow Region, Russia}.
  Springer, 2020, pp. 246--254.

\bibitem{liang2022ecg_segnet}
Xiaohong Liang, Liping Li, Yuanyuan Liu, Dan Chen, Xinpei Wang, Shunbo Hu,
  Jikuo Wang, Huan Zhang, Chengfa Sun, and Changchun Liu,
\newblock ``Ecg\_segnet: An ecg delineation model based on the encoder-decoder
  structure,''
\newblock {\em Computers in Biology and Medicine}, vol. 145, pp. 105445, 2022.

\bibitem{peimankar2021dens}
Abdolrahman Peimankar and Sadasivan Puthusserypady,
\newblock ``Dens-ecg: A deep learning approach for ecg signal delineation,''
\newblock {\em Expert systems with applications}, vol. 165, pp. 113911, 2021.

\bibitem{abrishami2018supervised}
Hedayat Abrishami, Chia Han, Xuefu Zhou, Matthew Campbell, and Richard Czosek,
\newblock ``Supervised ecg interval segmentation using lstm neural network,''
\newblock in {\em Proceedings of the International Conference on Bioinformatics
  \& Computational Biology (BIOCOMP)}. The Steering Committee of The World
  Congress in Computer Science, Computer~…, 2018, pp. 71--77.

\bibitem{wang2020knowledge}
Jilong Wang, Renfa Li, Rui Li, and Bin Fu,
\newblock ``A knowledge-based deep learning method for ecg signal
  delineation,''
\newblock {\em Future Generation Computer Systems}, vol. 109, pp. 56--66, 2020.

\bibitem{avetisyan2024deep}
Aram Avetisyan, Shahane Tigranyan, Ariana Asatryan, Olga Mashkova, Sergey
  Skorik, Vladislav Ananev, and Yury Markin,
\newblock ``Deep neural networks generalization and fine-tuning for 12-lead ecg
  classification,''
\newblock {\em Biomedical Signal Processing and Control}, vol. 93, pp. 106160,
  2024.

\bibitem{ribeiro2020automatic}
Ant{\^o}nio~H Ribeiro, Manoel~Horta Ribeiro, Gabriela~MM Paix{\~a}o, Derick~M
  Oliveira, Paulo~R Gomes, J{\'e}ssica~A Canazart, Milton~PS Ferreira, Carl~R
  Andersson, Peter~W Macfarlane, Wagner Meira~Jr, et~al.,
\newblock ``Automatic diagnosis of the 12-lead ecg using a deep neural
  network,''
\newblock {\em Nature communications}, vol. 11, no. 1, pp. 1760, 2020.

\bibitem{wagner2020ptb}
Patrick Wagner, Nils Strodthoff, Ralf-Dieter Bousseljot, Dieter Kreiseler,
  Fatima~I Lunze, Wojciech Samek, and Tobias Schaeffter,
\newblock ``Ptb-xl, a large publicly available electrocardiography dataset,''
\newblock {\em Scientific data}, vol. 7, no. 1, pp. 154, 2020.

\bibitem{howard2017mobilenets}
Andrew~G Howard, Menglong Zhu, Bo~Chen, Dmitry Kalenichenko, Weijun Wang,
  Tobias Weyand, Marco Andreetto, and Hartwig Adam,
\newblock ``Mobilenets: Efficient convolutional neural networks for mobile
  vision applications,''
\newblock {\em arXiv preprint arXiv:1704.04861}, 2017.

\end{thebibliography}
\end{document}